\documentclass[letterpaper, 10pt, conference, twoside]{style/ieeeconf}    % Use this line for a4 paper
\makeatletter
\let\NAT@parse\undefined
\makeatother

\usepackage{dblfloatfix}

\usepackage[numbers,sectionbib,sort&compress]{natbib}
\usepackage{bm}
\usepackage{caption}
\usepackage{gensymb}
\usepackage{graphicx}
\usepackage{amsmath}
\usepackage{amssymb}
\usepackage{subcaption}
\usepackage{amsfonts}
\usepackage{siunitx}
\usepackage{booktabs}
\usepackage{makecell}
\usepackage{multirow}
\usepackage{upgreek}
\usepackage[font=small]{caption}
\usepackage[export]{adjustbox}
\usepackage{tikz}
\usepackage{tabularx}
\usepackage{svg}
\usepackage{sidecap} \sidecaptionvpos{figure}{c}
\newcommand\edit[1]{{\color{black}#1}}

\usepackage{hyperref}
\hypersetup{colorlinks, breaklinks, linkcolor=[rgb]{0.,0.,0.}, citecolor=[rgb]{0.000,0.427,0.173}, urlcolor=[rgb]{0.031,0.318,0.612}}

\usepackage[nameinlink, capitalize]{cleveref}
\usepackage{color}
\definecolor{CommentPink}{rgb}{1,0.2,0.5}
\definecolor{CommentBlue}{rgb}{0,0,1}
\definecolor{CommentGreen}{rgb}{0,1,0}

\Crefname{section}{Sec.}{Sec.}
\Crefname{equation}{Eq.}{Eq.}

\DeclareMathOperator{\Tr}{Tr}

\usepackage[printonlyused,withpage,nolist,nohyperlinks]{acronym}
\begin{acronym}

\acro{2D}{two-dimensional}

\acro{3D}{three-dimensional}

\acro{AHRS}{attitude and heading reference system}

\acro{AUV}{autonomous underwater vehicle}
\acro{AGV}{autonomous ground vehicle}

\acro{CPP}{Chinese Postman Problem}

\acro{DoF}{degree of freedom}
\acrodefplural{DoF}[DoFs]{degrees of freedom}

\acro{DVL}{Doppler velocity log}

\acro{FSM}{finite state machine}

\acro{IMU}{inertial measurement unit}

\acro{LBL}{Long Baseline}

\acro{MCM}{mine countermeasures}

\acro{MDP}{Markov decision process}
\acrodefplural{MDP}[MDPs]{Markov decision processes}

\acro{POMDP}{Partially Observable Markov Decision Process}
\acrodefplural{POMDP}[POMDPs]{Partially Observable Markov Decision Processes}

\acro{PRM}{Probabilistic Roadmap}
\acrodefplural{PRM}[PRM]{Probabilistic Roadmaps}

\acro{ROI}{region of interest}
\acrodefplural{ROI}[ROIs]{regions of interest}

\acro{ROS}{Robot Operating System}

\acro{ROV}{remotely operated vehicle}

\acro{RRT}{Rapidly-exploring Random Tree}
\acrodefplural{RRT}[RRTs]{Rapidly-exploring Random Trees}

\acro{SLAM}{Simultaneous Localisation and Mapping}

\acro{SSE}{sum of squared errors}

\acro{STOMP}{Stochastic Trajectory Optimization for Motion Planning}

\acro{TRN}{Terrain-Relative Navigation}

\acro{UAV}{unmanned aerial vehicle}

\acro{USBL}{Ultra-Short Baseline}

\acro{IPP}{informative path planning}

\acro{FoV}{field of view}
\acrodefplural{FoV}[FoVs]{fields of view}

\acro{CDF}{cumulative distribution function}

\acro{ML}{maximum likelihood}

\acro{RMSE}{Root Mean Squared Error}
\acro{MLL}{Mean Log Loss}

\acro{GP}{Gaussian Process}
\acrodefplural{GP}[GPs]{Gaussian Processes}

\acro{KF}{Kalman Filter}

\acro{IP}{Interior Point}
\acro{BO}{Bayesian Optimization}

\acro{SE}{squared exponential}

\acro{UI}{uncertain input}

\acro{MCL}{Monte Carlo Localisation}
\acro{AMCL}{Adaptive Monte Carlo Localisation}

\acro{SSIM}{Structural Similarity Index}

\acro{MAE}{Mean Absolute Error}
\acro{RMSE}{Root Mean Squared Error}
\acro{AUSE}{Area Under the Sparsification Error curve}

\acro{AL}{active learning}
\acro{DL}{deep learning}
\acro{CNN}{convolutional neural network}

\acro{MC}{Monte-Carlo}

\acro{GSD}{ground sample distance}

\acro{BALD}{Bayesian active learning by disagreement}

\acro{fCNN}{fully convolutional neural network}
\acro{FCN}{fully convolutional neural network}

\acro{CMA-ES}{covariance matrix adaptation evolution strategy}

\acro{FoV}{field of view}

\acro{mIoU}{mean Intersection-over-Union}
\acro{ECE}{expected calibration error}

\acro{GAN}{Generative Adversarial Network}
\acro{POMCP}{Partially Observable Monte-Carlo Planning}
\acro{MCTS}{Monte-Carlo tree search}
\acro{RL}{reinforcement learning}

\end{acronym}

\IEEEoverridecommandlockouts                           % This command is only needed if
\title{Deep Reinforcement Learning with Dynamic Graphs \\ for Adaptive Informative Path Planning}
\author{Apoorva Vashisth$^{1}$ \and Julius R\"{u}ckin$^{2}$ \and Federico Magistri$^{2}$ \and Cyrill Stachniss$^{2}$ \and Marija Popovi\'{c}$^{2}$
\thanks{$^1$A.V. is with the Department of Mechanical Engineering, Indian Institute of Technology, Kharagpur, India. $^2$J.R., F.M., C.S., and M.P. are with the Institute of Geodesy and Geoinformation, Cluster of Excellence PhenoRob, University of Bonn. M.P. is also with MAVLab, Faculty of Aerospace Engineering, TU Delft. C.S. is also with the University of Oxford and Lamarr Institute for Machine Learning and Artificial Intelligence, Germany. This work has been funded by the Deutsche Forschungsgemeinschaft (DFG, German Research Foundation) under Germany's Excellence Strategy, EXC-2070 -- 390732324 (PhenoRob). Corresponding: \texttt{m.popovic@tudelft.nl}.}%}
}

\begin{document}

\maketitle

%%%%%%%%%%%%%%%%%%%%%%%%%%%%%%%%%%%%%%%%%%%%%%%%%%%%%%%%%%%%%%%%%%%%%%%%%%%%%%%%
\begin{abstract}

Autonomous robots are often employed for data collection due to their efficiency and low labour costs. A key task in robotic data acquisition is planning paths through an initially unknown environment to collect observations given platform-specific resource constraints, such as limited battery life. Adaptive online path planning in 3D environments is challenging due to the large set of valid actions and the presence of unknown occlusions. To address these issues, we propose a novel deep reinforcement learning approach for adaptively replanning robot paths to map targets of interest in unknown 3D environments. A key aspect of our approach is a dynamically constructed graph that restricts planning actions local to the robot, allowing us to react to newly discovered \edit{static} obstacles and targets of interest. For replanning, we propose a new reward function that balances between exploring the unknown environment and exploiting online-\edit{discovered} targets of interest. Our experiments show that our method enables more efficient target \edit{discovery} compared to state-of-the-art learning and non-learning baselines. We also showcase our approach for orchard monitoring using an unmanned aerial vehicle in a photorealistic simulator. \edit{We open-source our code and model at: \url{https://github.com/dmar-bonn/ipp-rl-3d}.}

\end{abstract}

\acresetall

%%%%%%%%%%%%%%%%%%%%%%%%%%%%%%%%%%%%%%%%%%%%%%%%%%%%%%%%%%%%%%%%%%%%%%%%%%%%%%%%
\section{Introduction}

Efficient data collection is a key requirement in many monitoring tasks, such as environmental mapping~\citep{dhariwal2004bacterium, duarte2016application, marchant2012bayesian, ruckin2023informative, wang2023spatio}, precision agriculture~\citep{magistri2020segmentation, carbone2022monitoring, magistri2019using}, and exploration~\citep{binney2010informative, hitz2017adaptive, cao2023ariadne}.  Autonomous robots are becoming popular tools for mobile sensing applications since they offer labour- and cost-effective alternatives to using conventional platforms%~\citep{fernandez2018current}
, manual approaches~\citep{su2022ai}, or static 
sensing methods~\citep{al2013efficient}. A key challenge in this context is planning paths that maximise the information value of collected data in large environments with limited onboard resources, e.g. mission time or battery capacity.

In this work, our goal is to map a set of targets of interest in an initially unknown 3D environment using an \ac{UAV} with a unidirectional sensor as efficiently as possible. Possible applications for such a system are finding apples in an orchard, victims in a search and rescue scenario, or components in a warehouse. We cast this problem as the \textit{informative path planning} problem, which aims to maximise the information value of obtained sensor observations subject to resource constraints, e.g. maximum path length or battery capacity. Our problem considers \textit{adaptively} replanning robot paths to account for observations collected online. In our setting, adaptive replanning is challenging due to unknown view-dependent occlusions in the 4D action space, i.e. the \ac{UAV} 3D position and yaw angle.

\begin{figure}[!t]
\centering
% \vspace{2mm}
\includegraphics[width=\columnwidth]{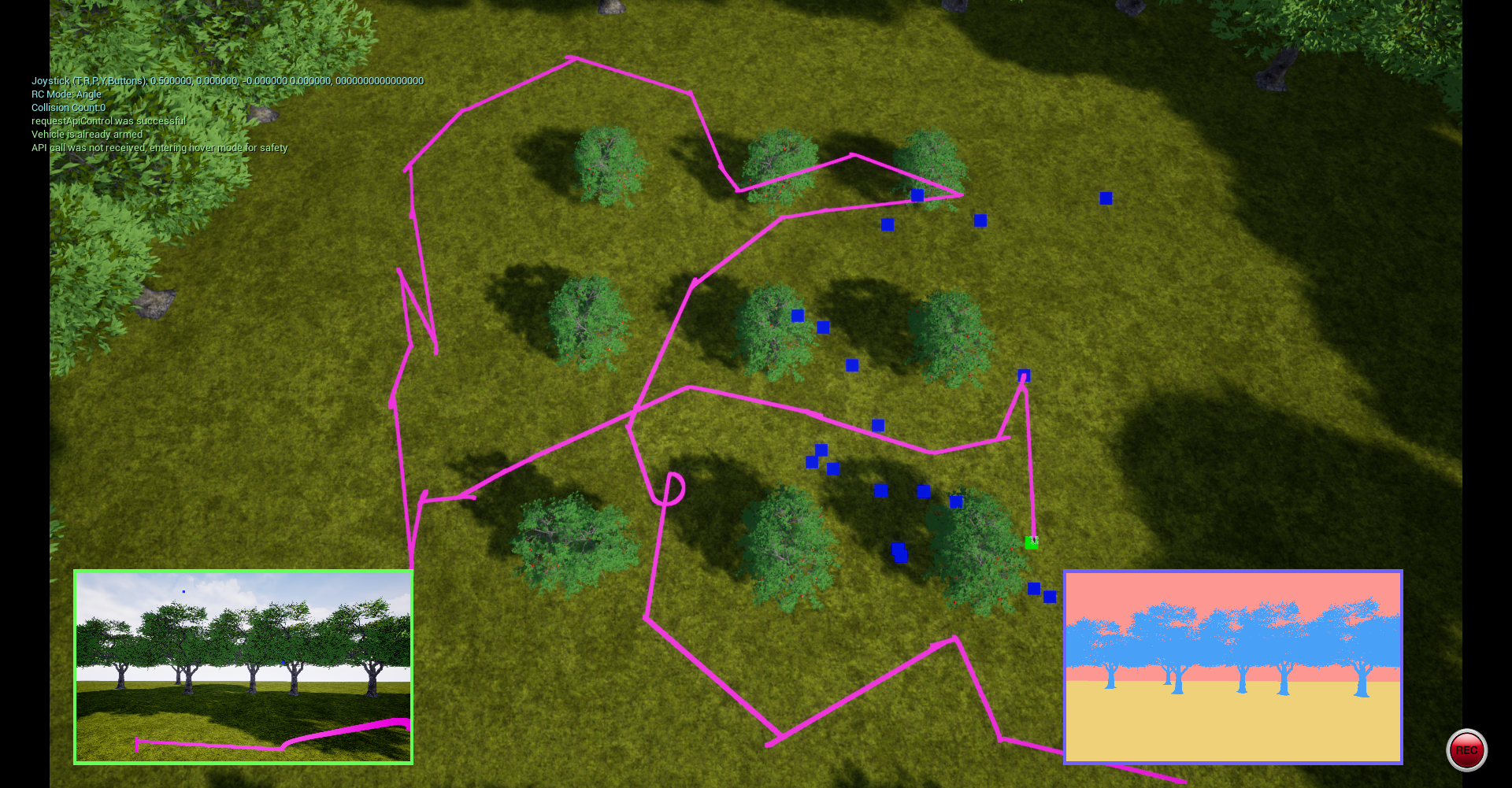}
% \vspace{-0.0cm}
\caption{Our reinforcement learning approach for adaptive informative path planning applied in an orchard monitoring scenario using an unmanned aerial vehicle (UAV).  Blue squares are candidate waypoints output by our planner, while the green square is the chosen next waypoint to visit. The inset windows show the onboard camera view and semantic segmentation for \edit{discovering apples}. By planning collision-free paths for the UAV online, we maximise the number of apple fruits discovered under flight-\edit{length} constraints.
}\small
\label{F:teaser}
% \vspace{-0.6cm}
\end{figure}

Classical approaches for this problem precompute a path before the mission starts~\citep{binney2012branch, arora2017randomized, binney2010informative, marchant2014bayesian} \edit{without considering online-collected observations for replanning, likely leading to sub-optimal performance}. In contrast, adaptive informative path planning approaches allow robots to replan their paths online based on newly collected data~\citep{hitz2017adaptive, ercolani2022clustering, popovic2020informative, lim2016adaptive, schmid2020efficient, mascarich2019distributed, zhu2021online, meera2019obstacle, rhodes2020informative, stache2023adaptive, ruckin2023informative, bouman2022adaptive, palazzolo2018effective, osswald2016speeding}. However, these methods are typically computationally inefficient in complex environments involving high-dimensional action spaces, such as \ac{UAV}-based applications. Recently, approaches using deep reinforcement learning have been proposed for adaptive informative planning that outperform non-learning methods in various scenarios~\citep{wei2020informative, cao2023catnipp, choi2021adaptive, ruckin2022adaptive}. These methods are not directly applicable to our problem setting as they do not adjust the sensor orientation~\citep{choi2021adaptive, cao2023catnipp, wei2020informative} or assume obstacle-free workspaces~\citep{ruckin2022adaptive,cao2023catnipp}. Extending these approaches for obstacle avoidance is non-trivial, requiring updating the action space as the obstacles are discovered. %For example, the probabilistic roadmaps (PRM)~\citep{kavraki1996probabilistic} utilised by \citet{cao2023catnipp} need to be completely rebuilt by removing invalid paths and waypoints to account for detected obstacles.

The main contribution of this paper is a deep reinforcement learning-based approach for adaptive replanning in unknown 3D environments to maximise the discovered targets of interest. A key aspect of our approach is a dynamically constructed graph representing the action space, which supports sequentially selecting the next best waypoint for the robot to visit. Our dynamic graph restricts planning to actions in the robot's local region at each timestep of the mission. In contrast to previous works~\citep{cao2023catnipp, ercolani2022clustering, hitz2017adaptive} reasoning about static predefined action spaces, this enables us to plan informative collision-free paths, even without prior knowledge about the environment. Combining our dynamic graph with sequential decision-making through reinforcement learning allows us to plan long-horizon paths. To capture the adaptive informative planning objective, we propose a new reward function that encourages the robot to both explore unknown regions and exploit \edit{discovered} targets. 
%We evaluate the percentage of targets \edit{discovered} during a mission, showing that our approach outperforms baselines. 
\cref{F:teaser} exemplifies our approach applied on a \ac{UAV} to discover fruits in an orchard.

In sum, we make the following three claims.
First, our approach enables more efficiently \edit{discovering} targets of interest compared to state-of-the-art non-learning-based and learning-based planning methods, including in previously unseen environments.
Second, by adapting the action space online, our dynamic graph ensures collision-free navigation in initially unknown environments while performing on par with or outperforming static global representations.
Third, our proposed reward function outperforms using purely exploratory rewards.
We validate the performance of our approach in a realistic orchard monitoring \ac{UAV} application.
%We open-source our code and provide a pre-trained model for the community at: \url{https://github.com/dmar-bonn/ipp-rl-3d}.
% \vspace{-0.2cm}

%%%%%%%%%%%%%%%%%%%%%%%%%%%%%%%%%%%%%%%%%%%%%%%%%%%%%%%%%%%%%%%%%%%%%%%%%%%%%%%%
\section{Related Work}
\label{relWork}

Informative path planning approaches are extensively applied in autonomous exploration and monitoring tasks. Classical methods~\citep{binney2012branch, arora2017randomized, binney2010informative, marchant2014bayesian} either plan a path offline or optimise paths to cover the complete robot workspace~\citep{jing2019coverage}. These combinatorial methods do not allow for online replanning due to the large computational burden incurred when exhaustively evaluating all possible paths through the environment. 
%Hence, these approaches are not applicable to our problem setting.

Generally, computational complexity is reduced by discretising the continuous action space by sampling and connecting candidate waypoints through platform-dependent paths. The robot visits only the sampled waypoints, traversing predefined paths formed by static global graphs representing the entire environment. Recent approaches~\citep{hollinger2014sampling, wang2019semantic, moon2022tigris} incrementally build the graph based on the current robot pose, reducing the replanning time. However, like classical approaches, these methods are non-adaptive to already collected observations for online replanning during a mission.

Adaptive informative path planning approaches~\citep{hitz2017adaptive, lim2016adaptive, schmid2020efficient, ercolani2022clustering, mascarich2019distributed, zhu2021online, meera2019obstacle, rhodes2020informative, stache2023adaptive, ruckin2023informative, popovic2020informative, bouman2022adaptive, palazzolo2018effective, osswald2016speeding} % %popovic2017online
replan robot paths online and consider gathered observations to inform subsequent decision steps. Several studies apply evolutionary algorithms to optimise paths for UAVs~\citep{meera2019obstacle, popovic2020informative, palazzolo2018effective} %popovic2017online,  
, autonomous surface vehicles~\citep{hitz2017adaptive}, \edit{or autonomous ground vehicles~\citep{bouman2022adaptive}} for adaptive replanning in a receding-horizon manner. \edit{\citet{meera2019obstacle} utilise covariance matrix adaptation evolution strategy (CMA-ES) to optimise the global path to map 2D target distributions using downward-facing cameras. \citet{bouman2022adaptive} maximise the environment coverage, planning in the robot's local region over 2D action spaces using omnidirectional sensors}. \citet{ercolani2022clustering} map gas distributions using nano aerial vehicles, separating path planning into global and local stages reasoning over clusters of sampled waypoints instead of individual waypoints. Similarly, \citet{lim2016adaptive} cluster waypoints by solving a group Steiner problem and frame UAV path planning as a travelling salesman problem over clusters. \citet{osswald2016speeding} combine globally optimal travelling salesman problem solutions on a coarse scale with effective local exploration adapting to the environment. %In contrast to our work, they consider obstacle-free workspaces. 
\citet{ruckin2023informative} and~\citet{mascarich2019distributed} derive information-theoretic planning objective to guide a  \ac{UAV} to cater for sensing uncertainty assuming obstacle-free workspaces. \citet{schmid2020efficient} propose new techniques for node rewiring in sampling-based path planning strategies utilising a point sensor. A major limitation of adaptive replanning methods is the computationally expensive online evaluation of information values of many candidate paths in complex environments.

\begin{figure*}[t]
\centering
\includegraphics[width=0.90\linewidth]{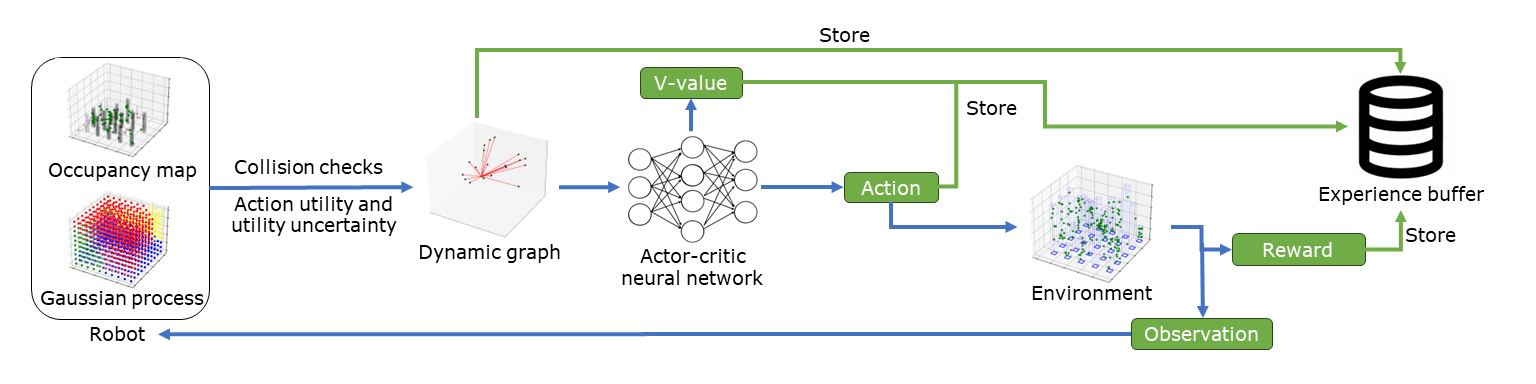}
\caption{At each mission timestep $t$, our approach samples collision-free waypoints in the robot's local environment. These waypoints, with considered yaw directions, generate action nodes. Each action node is associated with utility value and uncertainty of the utility value, regressed from the Gaussian process, to generate the dynamic graph. Our actor-critic network uses the dynamic graph to output the robot's state value and predicts the next action to execute, which generates a reward and observations from the environment. Blue arrows indicate the robot control loop and green indicate variables stored in the experience buffer to train the actor-critic network via on-policy learning.}\small
\label{F:overview}
% \vspace{-0.5cm}
\end{figure*}

Recent studies combine neural networks and reinforcement learning to solve the informative path planning problem~\citep{ruckin2022adaptive, choi2021adaptive, cao2023catnipp, zeng2022deep, wei2020informative}. Reinforcement learning-based solutions offer the benefits of computational efficiency at deployment and the ability to generalise to similar environments not seen during training. While \citet{ruckin2022adaptive} combine Monte Carlo tree search with a convolutional neural network, \citet{choi2021adaptive} consider advantages of paths planned by multiple low-level controllers. A further work by~\citet{cao2023catnipp} propose an attention-based neural network to achieve context-aware path planning in 2D workspaces, whereas~\citet{wei2020informative} use recurrent neural networks trained with Q-learning to plan the path. These methods are not directly applicable to our problem setup as they \edit{assume obstacle-free environments~\citep{ruckin2022adaptive, choi2021adaptive, cao2023catnipp, wei2020informative}, map 2D information distributions~\citep{ruckin2022adaptive, cao2023catnipp}, or do not account for UAV yaw~\citep{choi2021adaptive, cao2023catnipp, wei2020informative}}. Our approach is most similar to~\citet{cao2023catnipp}, which plans over a probabilistic roadmap-based environment representation in obstacle-free 2D workspaces. %Additionally, CAtNIPP rewards the actions contributing only to uncertainty reduction of the environment information distribution, whereas in our case the problem calls for a different definition of information. 
A key difference with respect to their work is \edit{our reinforcement learning-based framework restricting robot's actions to local area, enabling planning in an environment containing unknown obstacles.} Moreover, our proposed reward function not only encourages exploratory behaviour, as done in previous studies ~\citep{wei2020informative, cao2023catnipp}, but also incentivises exploiting newly collected information. We show that policies learned on our reward function outperform the ones learned on purely exploratory rewards.

%%%%%%%%%%%%%%%%%%%%%%%%%%%%%%%%%%%%%%%%%%%%%%%%%%%%%%%%%%%%%%%%%%%%%%%%%%%%%%%%
\section{Our approach}
% \vspace{-0.05cm}

We propose a novel deep reinforcement learning-based informative path planning approach for maximising the number of discovered targets of interest in unknown 3D environments. \cref{F:overview} overviews our method. A key aspect of our approach is a graph restricting planning to actions in the robot's \edit{known local workspace, ensuring collision-free action transitions} and allowing generation of collision-free paths. We call this environment representation a \textit{dynamic graph} as it evolves to account for newly gathered observations. We estimate each action's utility value in the current dynamic graph as the number of targets observed upon executing it and its corresponding uncertainty using a Gaussian process. At each timestep, our policy network outputs a probability distribution over the actions in the dynamic graph. We use the obtained observations to train the Gaussian process, update the occupancy map, and generate rewards reflecting the informative planning objective. We develop a new reward function that considers both the reduction in utility uncertainty and the number of observed targets. An experience buffer collects the robot's dynamic graph, sampled action, predicted state value, and reward over several training episodes to train the actor-critic network using on-policy reinforcement learning.

% We propose a novel deep \ac{RL}-based \ac{IPP} approach for maximising the number of detected targets of interest in cluttered 3D environments. \cref{F:overview} overviews our method. A key aspect of our approach is a fully connected graph constructed in the robot's local area at each timestep. We call this technique of representing the environment a \textit{dynamic graph} since the complete graph evolves to account for newly gathered observations. We define a utility term for each action as the number of targets observed upon execution of that action. We associate each candidate action present in the dynamic graph with a utility value and uncertainty value associated with the regressed utility value obtained from a Gaussian process. At each timestep, our learning algorithm outputs a probability distribution over the candidate action nodes of the dynamic graph. We use the obtained observations to update the robot's occupancy map, train the Gaussian process, and generate rewards based on our new reward structure that considers both the reduction in environment uncertainty and the observed number of targets. The robot's dynamic graph, reward, and outputs from the learning algorithm are collected over the training episode in an experience buffer for on-policy training.

\subsection{Environment Modelling}
\label{Sec:GP}

Our aim is to map the distribution of targets of interest in a 3D environment with unknown obstacles \edit{assumed to be static, non-moving}. We use Gaussian processes to model the view-dependent number of observed targets. We also maintain an occupancy map to plan collision-free paths. Our mission budget $B\in \mathbb{R}^+$ is defined as the maximum cost of the traversed path. We define the robot workspace $\mathcal{A}$ as a set of actions $\mathbf{a}_i=[x_i, y_i, z_i, d_i]^\top$, where $x_i, y_i, z_i\in\mathbb{R}$ are the robot 3D position coordinates within environment bounds and $d_t\in\mathcal{D}$ is the yaw of the unidirectional onboard sensor, e.g., an RGB-D camera. The robot takes observations at each distance interval $h$ as it travels. The set $\mathcal{D}$ denotes a user-defined set of possible robot yaw directions. During planning, at each mission timestep $t$, we plan over a set of $L$ candidate actions in the set $\mathcal{A}_t \subseteq \mathcal{A}, ~\vert \mathcal{A}_t \vert = L$. The candidate actions are sampled uniformly at random from the $C$-neighbourhood around $\mathbf{a}_{t-1}$ \edit{within known free space and are checked for collision-free reachability along straight lines}, where $C$ is a constant specifying the extent of the robot's local region. \edit{In this work, we assume quasi-holonomic motion constraints for evaluating reachability of candidate actions. The reachability check can be adapted for non-holonomic motion constraints.}
% spatial distance. %The candidate actions are sampled uniformly at random and satisfy the constraint $\mathcal{A}_t = \{\mathbf{a}_{i,t} \in \mathcal{A} \,  \vert \, f(\mathbf{a}_{i,t}, \mathbf{a}_{t-1}) < C,  \,  i \in [0, L)\} \, ,$%
% \begin{equation}
%     \mathcal{R}_t = \{\mathbf{a}_t \in \mathcal{A} \,  \vert \, f(\mathbf{a}_t, \mathbf{a}_{t-1}) < C\} \, ,
% \label{E:dynaGr_eq}
% \end{equation}
%
%where $\mathbf{a}_{t-1}$ is the previously executed action, $f: \mathcal{A}\times\mathcal{A}\rightarrow\mathbb{R}^+$ is a function defining the spatial distance between any two actions, and $C\in \mathbb{R}^+$ is a user-defined constant specifying the extent of robot's local region.%that specifies the size of the volume considered local to the robot. 

After executing an action $\mathbf{a}_t$, the robot observes a certain number of targets. \edit{Each action $\mathbf{a}_t \in \mathcal{A}$ in the robot workspace is associated with its utility value $u(\mathbf{a}_t) \in \mathbb{R}^+$ reflecting the number of targets observed}. The number of observed targets is normalised by the total number of targets in the environment for stable policy training.
% After executing an action $\mathbf{a}_t$, the robot observes a certain number of targets. We define the utility function $u:\mathcal{A}\rightarrow\mathbb{R}^+$ as mapping action $\mathbf{a}_t$ to the observed number of targets. The observed targets for $u$ are normalised by the total number of targets in the environment to ensure normalised neural network input features for stable policy training. 

%We utilise a Gaussian process to infer the relation between an action vector $\mathbf{a}_t$ and its associated utility value $u(\mathbf{a}_t)$.

\edit{As the utility function $u:\mathcal{A}\rightarrow\mathbb{R}^+$ is (partially) unknown for actions $\mathbf{a}_t$, we need to estimate it. To this end, we utilise Gaussian processes widely used to estimate spatially correlated phenomena~\citep{Rasmussen2006, wei2020informative, hitz2017adaptive, cao2023catnipp}. We train a Gaussian process on the utility values of the actions executed in the past and exploit its utility estimates to inform the planning policy. The estimated variance allows our policy to consider the uncertainty over estimated utilities during planning.}

% Gaussian processes are widely used to represent spatially correlated phenomena~\citep{Rasmussen2006, wei2020informative, hitz2017adaptive, cao2023catnipp}. We use a Gaussian process to predict the candidate actions' utility values and associated variances. The Gaussian process is trained on the utility values of the actions executed in the past. We exploit the Gaussian process to inform the policy. The variance allows our policy to consider the uncertainty in predicted utility values during planning.

A Gaussian process is characterised by a mean function $m(\mathbf{a}_i) \triangleq \mathbb{E}[u(\mathbf{a}_i)]$ and a covariance function $k(\mathbf{a}_i, \mathbf{a}_i^\prime) \triangleq \mathbb{E}[(u(\mathbf{a}_i) - m(\mathbf{a}_i))(u(\mathbf{a}_i^\prime) - m(\mathbf{a}_i^\prime))]$ as $u(\mathbf{a}_i) \sim GP(m(\mathbf{a}_i),k(\mathbf{a}_i,\mathbf{a}_i^\prime))$, where $\mathbb{E}[\cdot]$ is the expectation operator and $\mathbf{a}_i, \mathbf{a}_i^\prime \in \mathcal{A}$. Hence, considering a set of candidate actions $\mathcal{A}_t$ at timestep $t$ for which we wish to infer the utility, let the actions in set $\mathcal{A}_t$ correspond to a feature matrix $\mathbf{A}_t$, where each $i^{th}$ row corresponds to the action vector $\mathbf{a}_t^i\in\mathcal{A}_t$. The set of all previously executed actions $\{\mathbf{a}_0, \mathbf{a}_2, \ldots, \mathbf{a}_{t-1}\}$ is represented by feature matrix $\mathbf{A}_{t-1}^\prime$. Utility values corresponding to previously executed actions are represented by the vector $\mathbf{u}_{t-1}^\prime = [u(\mathbf{a}_1), \ldots, u(\mathbf{a}_{t-1})]^\top $. We predict the utility values of candidate action set $\mathcal{A}_t$ by conditioning the Gaussian process on the observed utility values $\mathbf{u}_t \mid \mathbf{A}_{t-1}^\prime, \mathbf{u}_{t-1}^\prime, \mathbf{A}_t \sim \mathcal{N}(\boldsymbol{\upmu}, \mathbf{P})$:
\begin{align}
 \boldsymbol{\upmu}(\mathbf{A}_t, \mathbf{A}_{t-1}^\prime) ={}& \mathbf{m}(\mathbf{A}_t) + \mathbf{K}(\mathbf{A}_t,\mathbf{A}_{t-1}^\prime)\,[\mathbf{K}(\mathbf{A}_{t-1}^\prime,\mathbf{A}_{t-1}^\prime) &\notag\\
 & + \sigma_n^2\mathbf{I}]^{-1} 
  \big(\mathbf{u}_{t-1}^\prime - \mathbf{m}(\mathbf{A}_{t-1}^\prime)\big) \, \textrm{,} \label{E:gp_mean} &\\
 \mathbf{P}(\mathbf{A}_t, \mathbf{A}_{t-1}^\prime) ={}& \mathbf{K}(\mathbf{A}_t,\mathbf{A}_t) - \mathbf{K}(\mathbf{A}_t,\mathbf{A}_{t-1}^\prime)  [\mathbf{K}(\mathbf{A}_{t-1}^\prime,
 &\notag\\
 & \mathbf{A}_{t-1}^\prime) +  
  \sigma_n^2\mathbf{I}]^{-1} \mathbf{K}(\mathbf{A}_t,\mathbf{A}_{t-1}^\prime)^\top \, \textrm{,} \label{E:gp_cov}
\end{align}
where $\sigma_{n}^2 \in \mathbb{R}^+$ is a hyperparameter describing the measurement noise variance, $\mathbf{I}$ is an $n\times n$ identity matrix where $n=t-1$, and $\mathbf{K}(\cdot,\cdot)$ corresponds to the covariance matrix.

\subsection{Adaptive Informative Path Planning}
We model the path followed by the robot as a sequence of consecutively executed actions $\psi_0^T = (\mathbf{a}_0,\mathbf{a}_1,\ldots,\mathbf{a}_T)$ where $\mathbf{a}_0$ denotes the start action, i.e. the initial robot pose, and $\mathbf{a}_T$ is the final action upon depletion of the budget $B$. The general informative path planning problem aims to find an optimal path ${\psi^\ast}_0^{T} \in \Psi$ in the space of all possible paths $\Psi$ to optimise an information-theoretic objective function:
\begin{equation}
    {\psi^\ast}_0^{T} = \mathop{{\rm {argmax}}} \limits_{\psi \in \Psi}~{\rm \mathit{I}}(\psi),\ {\rm {s.t.}} \,\,  C(\psi) \leq B \, ,
    \label{eq:objective}
\end{equation}
where ${\rm \mathit{I}}:\psi_0^T \to \mathbb{R}^+$ is the information gained from observations obtained along path $\psi_0^T$, the cost function $C~:~\psi_0^T \to \mathbb{R}^+$ maps path $\psi_0^T$ to its execution cost.

Our robot traverses a straight line between two consecutive actions. Observations are equidistantly collected along the path $\psi_0^T$ at a frequency $h$ and are used to update the Gaussian process and generate a reward. Hence, we model the informative path planning problem as a sequential decision-making process. As we aim to maximise the number of observed targets, we define a function $\nu:\mathcal{A}\times\Psi\rightarrow\mathbb{R}^+$ as the number of new targets observed upon executing an action $\mathbf{a}_t\in\mathcal{A}$ after following the path $\psi_0^{t-1}$. Note that information $\nu(\mathbf{a}_t, \psi_0^{t-1})$ and utility $u(\mathbf{a}_t)$ differ as the utility measures the number of all targets observed upon executing an action, whereas information considers only targets that were newly observed after executing the action. Hence, modelling $\nu(\mathbf{a}_t, \psi_0^{t-1})$ with a Gaussian process would require including the temporal variations of an action's utility value, increasing the Gaussian process and policy training complexity. We therefore choose to model utility with a Gaussian process, as it only depends on a single action $\mathbf{a}_t$.

We define the information obtained along a path as:
\begin{equation}
    \mathit{I}(\psi_0^{T}) = \sum_{t=1}^{T} \, \nu(\mathbf{a}_t, \psi_{0}^{t-1}) \,,
\end{equation}
where we aim to plan a path $\psi_0^T$ to maximise information $\mathit{I}$.

For informative planning, we leverage our Gaussian process defined in \cref{Sec:GP} to regress the utility and uncertainty associated with actions.  We apply an upper confidence bound to the set of candidate actions $\mathcal{A}_{t}$ to obtain a subset of high-interest actions $\hat{\mathcal{A}_{t}}$ used in our reward function:
\begin{equation}
    \hat{\mathcal{A}_{t}}=\{\mathbf{a}_{i,t}\in \mathcal{A}_t\,|\, m(\mathbf{a}_{i,t}) +\beta k(\mathbf{a}_{i,t}, \mathbf{a}_{i,t})\geq \mu_{th}\}\,,    
\label{eq:ucb}
\end{equation}
where $m(\mathbf{a}_{i,t})$ and $k(\mathbf{a}_{i,t}, \mathbf{a}_{i,t})$ are the mean utility of action~$\mathbf{a}_{i,t}$ and corresponding variance inferred from the Gaussian process. The parameter $\beta \in \mathbb{R}$ controls the confidence interval width and $\mu_{th}$ is a user-defined threshold.

We introduce a new reward function that balances exploring the environment and exploiting collected information. The information criteria in previous works~\cite {cao2023catnipp, wei2020informative} consider environment exploration only. However, our problem considers discovering targets, therefore requiring a measure of information value in the reward. At each timestep $t$, the robot executes action $\mathbf{a}_t$, collects observations and receives a reward $r_t\in \mathbb{R}^+$. Our reward function consists of an exploration term $r_{e,t}$ and an information term $r_{u,t}$:
%
% \begin{align}
% \begin{split}
% \label{reward}
%     r_t(\mathbf{A}_t, \mathbf{A}^{t-1}, \mathbf{A}^{t}, \mathbf{a}_t, \psi_0^{t-1}) =& \alpha r_{e,t}(\mathbf{A}_t, \mathbf{A}^{t-1}, \mathbf{A}^{t}) + \\
%     &\delta r_{u,t}(\mathbf{a}_t, \psi_0^{t-1})\, ,\\
%     r_{e,t}(\mathbf{A}_t, \mathbf{A}^{t-1}, \mathbf{A}^{t}) =& \frac{\Tr(P(\mathbf{A}_t, \mathbf{A}^{t-1})) - \Tr(P(\mathbf{A}_t, \mathbf{A}^t))}{\Tr(P(\mathbf{A}_t, \mathbf{A}^{t-1}))} \, , \\
%     r_{u,t}(\mathbf{a}_t, \psi_0^{t-1}) &= \nu(\mathbf{a}_t, \psi_0^{t-1}) \, , \\
% \end{split}
% \end{align}
%

%
\begin{align}
\begin{split} \label{reward}
    r_t(\hat{\mathbf{A}}_t, \mathbf{A}_{t-1}^\prime, \mathbf{A}_{t}^\prime, \mathbf{a}_t, \psi_0^{t-1}) &= \\ \, 
    \alpha r_{e,t}(\hat{\mathbf{A}}_t, \mathbf{A}_{t-1}^\prime, &\mathbf{A}_{t}^\prime) + \, \beta r_{u,t}(\mathbf{a}_t, \psi_0^{t-1})\, ,\\
\end{split}
\end{align}
with:
\begin{align}
\begin{split}
    % r_t(\hat{\mathbf{A}}_t, \mathbf{A}_{t-1}^\prime, \mathbf{A}_{t}^\prime, \mathbf{a}_t, \psi_0^{t-1}) =& \, \alpha r_{e,t}(\hat{\mathbf{A}}_t, \mathbf{A}_{t-1}^\prime, \mathbf{A}_{t}^\prime) + \\
    % &\, \delta r_{u,t}(\mathbf{a}_t, \psi_0^{t-1})\, ,\\
    r_{e,t}(\hat{\mathbf{A}}_t, \mathbf{A}_{t-1}^\prime, \mathbf{A}_{t}^\prime) =& \frac{\Tr(\mathbf{P}(\hat{\mathbf{A}}_t, \mathbf{A}_{t-1}^\prime)) - \Tr(\mathbf{P}(\hat{\mathbf{A}}_t, \mathbf{A}_t^\prime))}{\Tr(\mathbf{P}(\hat{\mathbf{A}}_t, \mathbf{A}_{t-1}^\prime))} \, , \\
    r_{u,t}(\mathbf{a}_t, \psi_0^{t-1}) =& ~\nu(\mathbf{a}_t, \psi_0^{t-1}) \, , \\
\end{split}
\end{align}
where $\Tr(\cdot)$ is the trace operator of a matrix and $\alpha$ and $\beta$ are constants used to trade off exploration and exploitation.

The variance reduction of the Gaussian process measures exploration $r_{e,t}$. To this end, we maximise the decrease in the covariance matrix trace following the A-optimal design criterion~\citep{sim2005global}.  Scaling the reward by $\Tr(P(\hat{\mathbf{A}}_t, \mathbf{A}_{t-1}^\prime))$ stabilises the actor-critic network training~\citep{cao2023catnipp}. The term $r_{u,t}$ measures the new information gained after executing $\mathbf{a}_t$.

%We compute the utility variance after re-training the Gaussian process $GP(m(\mathbf{a}_t), k(\mathbf{a}_t, \mathbf{a}_t))$ on all obtained utility values $\mathbf{u}_{t}^\prime$

\subsection{Dynamic Graph Representation}

Adaptive informative path planning requires reasoning about the information distribution in the environment. We propose a dynamic graph that models the \edit{collision-free reachable} action space and information distribution in the robot's neighbourhood by sampling actions as defined in \cref{Sec:GP}, as opposed to a static global non-obstacle-aware representation~\citep{cao2023catnipp, ercolani2022clustering, hitz2017adaptive}. Our robot's policy relies on the representation of the current knowledge about the environment in the dynamic graph to predict next action~\cref{SS:rl_network}.
%Our dynamic graph efficiently represents the current knowledge about the environment, which the robot's policy uses to predict the next action (\cref{SS:rl_network}).

At each timestep $t$, we (re-)build a fully connected graph $\mathcal{G}_t=(\mathcal{N}_t, \mathcal{E}_t)$, where the node set $\mathcal{N}_t$ is the set of candidate actions $\mathcal{A}_t$ defined in \cref{Sec:GP}, to account for newly gathered observations. We randomly sample $K$ candidate positions $[x_i, y_i, z_i]^\top$ within a $C$-neighbourhood of the current robot position and create $L = K\times \vert\mathcal{D}\vert$ nodes by associating each position with possible yaws $\mathcal{D}$.
%, where $\vert\mathcal{D}\vert$ denotes the cardinality of set $\mathcal{D}$. 
The edge set $\mathcal{E}_t$ connects each pair of actions such that $e_{i, j, t} = (i,j,c_{i,j})\in\mathcal{E}_t$, where the cost $c_{i,j}$ is the edge weight, $i, j \leq L$, and $i \neq j$. \edit{The edge set $\mathcal{E}_t$ and the cost $c_{i,j}$ are based on the robot's motion constraints, defining the path traversed from action $\mathbf{a}_i$ to $\mathbf{a}_j$. In this work, we consider a holonomic UAV travelling between two actions on a straight line, hence t}he cost $c_{i,j}$ is defined as the sum of the actions' Euclidean distance and a small constant cost $C_s$ if the robot yaw changes. 
%Building the graph $\mathcal{G}_t$ only in the robot's local region allows us to quickly update the action space to account for newly gathered observations. % This also enables efficiently planning collision-free paths 
%Since we always plan over local regions, we can plan collision-free paths in an initially unknown environment while retaining constant runtime at deployment. % compared to a global graph such as CAtNIPP~\citep{cao2023catnipp}, which plans over $K\geq800$ action nodes in absence of obstacles.% where several unknown obstacles may be present. 

To better inform the planning policy, we leverage our Gaussian process to create node features utilised in our actor-critic network. At each timestep $t$, the node feature matrix $\mathbf{M}_t$ of graph $\mathcal{G}_t$ consists of the robot's candidate actions and the mean and variance values queried from the Gaussian process. The $i^{\mathrm{th}}$ row of $\mathbf{M}_t$ relates to the $i^{\mathrm{th}}$ action:
\begin{equation}
    \mathbf{M}_t(i)= [\mathbf{a}_{i,t},  \, m(\mathbf{a}_{i,t}),  \, k(\mathbf{a}_{i,t}, \mathbf{a}_{i,t})] \, ,
\label{E:augGr}
\end{equation}
where $\mathbf{a}_{i,t} = [x_{i,t}, \,  y_{i,t},  \, z_{i,t},  \, d_{i,t}]^\top$, and $m(\mathbf{a}_{i,t})$ and $k(\mathbf{a}_{i,t}, \mathbf{a}_{i,t})$ are the regressed action's utility and variance.

\subsection{Actor-Critic Neural Network for Reinforcement Learning}
\label{SS:rl_network}
We exploit the dynamic graph to model collision-free actions for the planning policy to reason about and represent the current knowledge of the information distribution in the environment. As the Gaussian process only predicts the utility of greedily executing a single next action, we use reinforcement learning to train our policy for informative path planning over long-horizon paths.

We use an attention-based neural network to parameterise our stochastic planning policy~$\pi(\mathcal{G}_t, \psi_0^{t-1}, B_r, \mu_{th}) \in [0, 1]^L$ that predicts a probability distribution over all $L$ actions $\mathcal{A}_t$ based on the current dynamic graph $\mathcal{G}_t$, previously executed path $\psi_0^{t-1}$, remaining budget $B_r$, and the mean threshold $\mu_{th}$ defining actions of interest in \cref{eq:ucb}. We follow the network structure proposed by~\citet{cao2023catnipp} consisting of an encoder and a decoder module. The attention-based encoder learns the dependencies between actions in $\mathcal{G}_t$. We condition the learned actions' latent dependencies on a planning state consisting of previously executed actions $\psi_0^{t-1}$, the remaining budget $B_r$, and the threshold $\mu_{th}$. A budget mask filters out actions not reachable within the remaining budget. Based on the conditioned latent action dependencies, a decoder outputs a probabilistic policy reasoning over all actions in the dynamic graph. During training, the decoder also estimates the value function~$V(\mathcal{G}_t, \psi_0^{t-1}, B_r, \mu_{th})~\in~\mathbb{R}$ following the current policy~$a_t \sim \pi(\mathcal{G}_t, \psi_0^{t-1}, B_r, \mu_{th})$. The estimated values, sampled actions, dynamic graphs, planning states, and rewards are stored in the experience buffer utilised to train the policy with an on-policy actor-critic reinforcement algorithm. In this work, we use proximal policy optimisation~\citep{schulman2017proximal}. During deployment, at each time step $t$, we choose the most informative action from $\pi(\mathcal{G}_t, \psi_0^{t-1}, B_r, \mu_{th})$.
%
%\begin{equation}
 %   \mathbf{a}_t = \mathop{{\rm {argmax}}} \limits_{\mathbf{a}_{i,t} \in \mathcal{A}_t}{\pi(\mathcal{G}_t, \psi_0^{t-1}, B_r, \mu_{th})_i}\, .
%\end{equation}
%

%%%%%%%%%%%%%%%%%%%%%%%%%%%%%%%%%%%%%%%%%%%%%%%%%%%%%%%%%%%%%%%%%%%%%%%%%%%%%%%%
\section{Experimental Results}
We experimentally validate our three claims on the task of \ac{UAV}-based fruit monitoring in apple orchards. First, our approach enables more efficiently discovering targets of interest compared to non-learning baselines and learning-based methods. Second, our dynamic graph action space enables collision-free path planning in unknown environments while performing on par with state-of-the-art static global graph representations. Third, our new reward function more effectively manages the exploration-exploitation trade-off compared to using purely exploratory rewards, leading to more efficient targeted mapping. We demonstrate the performance of our approach in a realistic orchard simulation, showcasing its applicability for a practical monitoring task in a previously unseen environment.

\subsection{Experimental Setup}
\label{Sec:implement}
\begin{figure}[t]
     \centering
     \begin{subfigure}[b]{0.380\linewidth}
         \centering
         \includegraphics[width=\linewidth]{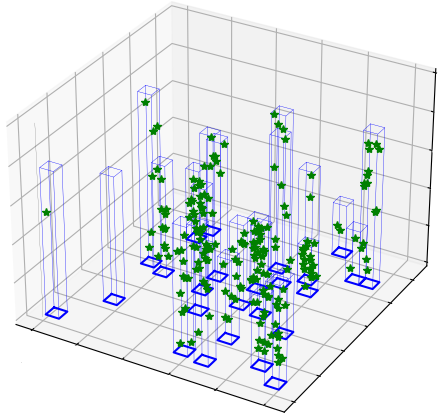}
        % \caption{ }
         \label{F:random}
     \end{subfigure}
     \hfill
     \begin{subfigure}[b]{0.394\linewidth}
         \centering
         \includegraphics[width=\linewidth]{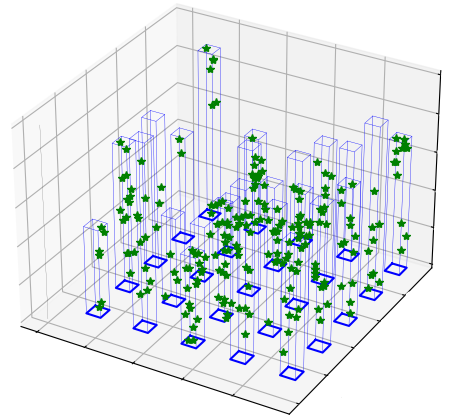}
       %  \caption{ }
         \label{F:grid}
     \end{subfigure}
     \hfill
      % \vspace{-0.2cm}
     \caption{Examples of testing and training orchard environments used in our experiments. \textit{Left}: Testing environment with trees placed at random locations. \textit{Right}: Training environment with trees placed in a regular square array. Thin and dark blue lines represent tree outlines and tree bases, respectively. Green stars indicate fruits.}\small
     \label{F:env}
\end{figure}

\textbf{Environment}. Our environment consists of trees and fruits bounded in a scale-agnostic unit cube. We maintain an occupancy map with $50 \times 50 \times 50$ voxels. During test phase, trees are generated at random positions in the environment but are arranged in a regularly spaced $5\times5$ array during training. \cref{F:env} shows examples of a training and a testing environment. In both cases, fruits are attached to generated trees at random positions. The occupancy grid map of the environment is initialised as unknown space and is updated via sensor observations with free space, observed fruits, and trees. For each observation, the utility value is used to train the Gaussian process detailed in \cref{Sec:GP}. We tune the hyperparameters in a small representative environment using the Matérn~$1/2$ kernel with $\mu_{th}=0.4$ and $\beta=1$ in~\cref{eq:ucb}. For our reward function defined in~\Cref{eq:objective}, we set $\alpha=1.0$ and $\delta=0.01$ to keep values of both terms numerically similar to balance between exploration and exploitation.

We consider a \ac{UAV} platform with an onboard RGB-D camera of $90^{\circ}$ \ac{FoV}. Since the confidence in discovered targets decreases with distance, we constrain the maximum detection range to 24$\%$ of the environment size. The \ac{UAV} can choose between yaw angles of $[0, \frac{\pi}{2}, \pi, \frac{3\pi}{2}]$ rad at the current altitude. Note that our method can be easily extended to finer discretisations.

\textbf{Training} An episode consists of a \ac{UAV} mission with a budget $B$. We train in a grid-based environment as a randomly initialised policy learns efficiently from this structure and transfers to randomised test environments. The total number of fruits varies between $200$ and $250$. We fix the number of positions $K$ in the dynamic graph and set the initial \ac{UAV} pose to $[0, 0, 0, \frac{\pi}{2}]^\top$. To keep our policy scale-agnostic, we normalise the robot's internal environment representation and action coordinates. Hence, the budget $B$ is unitless and randomly generated for each episode within the range $[7, 9]$. We fuse an observation into the occupancy map and Gaussian process each time the \ac{UAV} travels $0.2$ units from the position of the previous observation.

We terminate an episode if the maximum number of executed actions exceeds $256$. To speed up training, we run $12$ environments in parallel. The policy is trained for $8$ epochs using a batch size of $64$ and the Adam optimiser with a learning rate of $10^{-4}$, which decays by a factor of $0.96$ each $32$ optimisation steps. The policy gradient epsilon-clip parameter is set to $0.2$. Our model is trained on a workstation equipped with an Intel(R) Xeon(R) W-2133 CPU @ 3.60GHz and one NVIDIA Quadro RTX 5000 GPU. Our policy is trained for $\sim 440,000$ environment interactions.

\subsection{Comparison Against Baselines}
\label{Sec:BaseComp}
The first set of experiments shows that our dynamic graph-based reinforcement learning approach outperforms state-of-the-art baselines. We generate $25$ test environments corresponding to different random seeds and run $20$ trials on each environment instance with a budget of $10$ units.

\begin{figure}[t]
\centering
\includegraphics[width=0.80\linewidth]{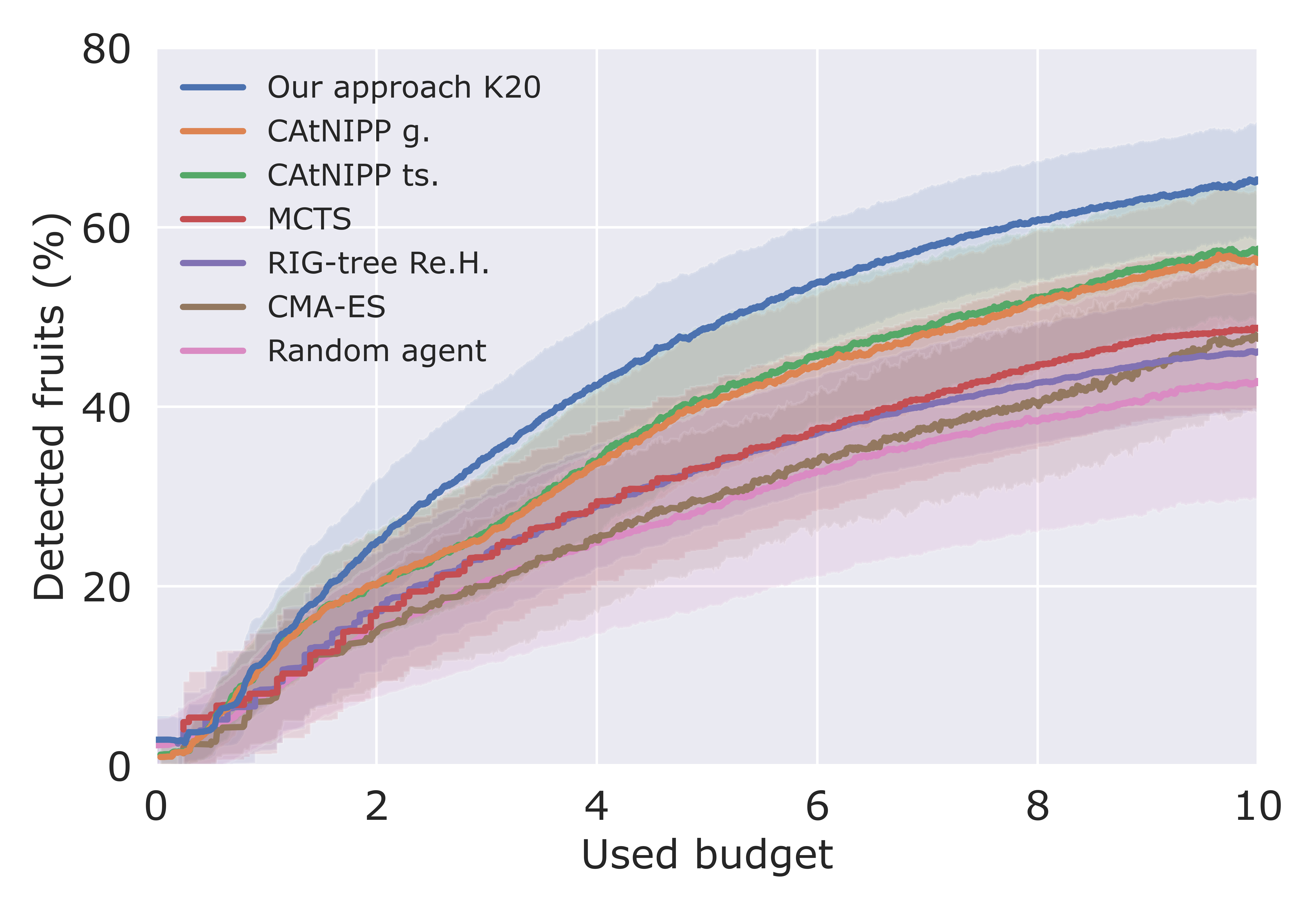}
% \vspace{-0.7cm}
\caption{Comparison of our approach against baselines in a \ac{UAV}-based fruit monitoring scenario. Solid lines indicate means over 500 trials and shaded regions show standard deviations. In our approach, using our exploration-exploitation reward function with a dynamic graph action space for reinforcement learning enables more efficiently discovering targets of interest (fruit) during a mission.
}\small
\label{F:baseComp}
% \vspace{-0.65cm}
\end{figure}

\begin{table}[!t]
    \vspace{0cm} \vfill
    \small
    \centering
    \caption{Comparison of our approach against baselines in a \ac{UAV}-based fruit monitoring scenario. Our dynamic graph-based reinforcement learning approach outperforms learning and non-learning baselines while maintaining low replanning time.}%The cells of the first column indicate mean values and standard deviation of the percentage detected fruits, and the second column indicates time taken per step in seconds.}
\begin{tabular}{l|p{1.75cm}<{\centering}|p{1cm}<{\centering}p{1cm}<{\centering}}
%    \toprule
         \multicolumn{1}{c|}{\multirow{1}{*}{Baseline}} & \multicolumn{1}{c|}{$\%$ targets} & \multicolumn{1}{c}{Time (s)} \\
         \midrule
         Our approach $K=20$ & $65.49 \pm 6.23$ & $1.58$\\ 
         CAtNIPP (ts.) & $57.14 \pm 7.60$ & $44.51$\\ 
         CAtNIPP (g.) & $56.49 \pm 7.65$ & $1.28$\\ 
         MCTS & $50.15 \pm 8.98$ & $218.83$ \\
         CMA-ES & $48.72 \pm 8.36$ & $165.41$ \\
         RIG-tree (Re.H.) & $47.43 \pm 6.51$ & $80.27$\\ 
         Random agent & $43.91 \pm 12.64$ & $0.28$\\ 
%         \bottomrule
    \end{tabular}
    \label{T:baseTable}
% \vspace{-0.35cm}
\end{table}

% \begin{table}[!t]
%     \vspace{0cm} \vfill
%     \small
%     \centering
%     \caption{Comparison of our approach against baselines in a \ac{UAV}-based fruit monitoring scenario. Our dynamic graph-based reinforcement learning approach outperforms learning and non-learning baselines while maintaining low replanning time.}%The cells of the first column indicate mean values and standard deviation of the percentage detected fruits, and the second column indicates time taken per step in seconds.}
% \begin{tabular}{l|p{1.75cm}<{\centering}|p{1.75cm}<{\centering}|p{1cm}<{\centering}p{1cm}<{\centering}}
% %    \toprule
%          \multicolumn{1}{c|}{\multirow{1}{*}{Baseline}} & \multicolumn{1}{c|}{$\%$ targets} & \multicolumn{1}{c|}{Double} & \multicolumn{1}{c}{Time (s)} \\
%          \midrule
%          Our approach & $65.49 \pm 6.23$ & $66.30 \pm 5.97$ & $1.58$\\ 
%          CAtNIPP (ts.) & $57.14 \pm 7.60$ & $57.56 \pm 6.76$ & $44.51$\\ 
%          CAtNIPP (g.) & $56.49 \pm 7.65$ & $55.95 \pm 6.91$ & $1.28$\\ 
%          MCTS & $50.15 \pm 8.98$ & $0.00 \pm 0.00$ & $218.83$ \\
%          CMA-ES & $48.72 \pm 8.36$ & $0.00 \pm 0.00$ & $165.41$ \\
%          RIG-tree (Re.H.) & $46.10 \pm 6.63$ & $0.00 \pm 0.00$ & $79.12$\\ 
%          Random agent & $43.91 \pm 12.64$ & $0.00 \pm 0.00$ & $0.28$\\ 
% %         \bottomrule
%     \end{tabular}
%     \label{T:baseTable}
% % \vspace{-0.35cm}
% \end{table}

For our approach, we consider $K=20$ sampled waypoints, $L=80$ nodes in the dynamic graph, and the reward function described in~\Cref{eq:objective}.  The learning-based baseline for evaluation is CAtNIPP~\citep{cao2023catnipp}, the state-of-the-art approach closest to our work which uses global graph-based planning. Since CAtNIPP considers obstacle-free environments and modifying its global graph to account for unknown obstacles is a non-trivial task, we allow the \ac{UAV} to pass through obstacles to ensure a fair comparison.

We compare against: (i)~CAtNIPP with a zero-shot policy (CAtNIPP g.) where the highest probability action is executed; (ii)~CAtNIPP with a trajectory sampling policy~\citep{cao2023catnipp} (CAtNIPP ts.) where four $8$-step paths are planned and the path with highest uncertainty reduction is executed for $3$ steps; (iii)~a random policy on $K=20$ dynamic graph construction (random agent); (iv)~non-learning rapidly exploring random information gathering trees~\citep{hollinger2014sampling} applied in a receding-horizon manner (RIG-tree Re.H.); \edit{(v)~non-learning Monte Carlo Tree Search~\citep{ott2023sequential} (MCTS); and (vi) non-learning Covariance Matrix Adaptation Evolution Strategy (CMA-ES)~\citep{popovic2017multiresolution}}.
To evaluate planning performance, we measure the percentage of discovered targets of interest (apples) during the test. We also report average replanning time per step.

\begin{figure}[t]
     \centering
     \begin{subfigure}[b]{0.38\linewidth}
         \centering
         \includegraphics[width=\linewidth]{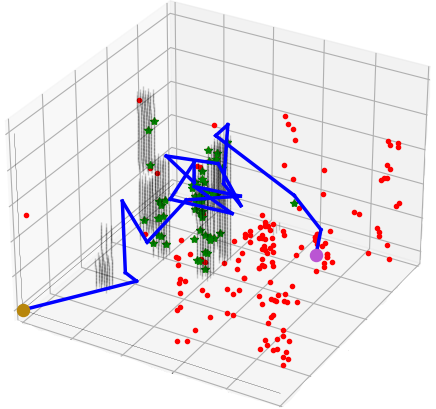}
         \caption{CAtNIPP (g.)}
         \label{F:catnipp}
     \end{subfigure}
     \hfill
     \begin{subfigure}[b]{0.38\linewidth}
         \centering
         \includegraphics[width=\linewidth]{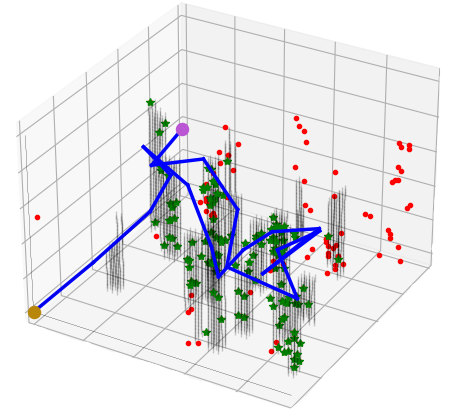}
         \caption{Our approach}
         \label{F:our_approach}
     \end{subfigure}
     \hfill
     \caption{Comparison of paths planned by (a) the global graph-based CAtNIPP~\citep{cao2023catnipp} baseline and (b) our dynamic graph-based reinforcement learning approach with $K=20$ in a fruit monitoring scenario. The blue line shows the executed \ac{UAV} path, with the brown and pink circles indicating start and end positions. Red dots are targets not yet observed, while green stars are observed targets.}\small
     \label{F:Qual_comp}
     % \vspace{-0.6cm}
\end{figure}

\cref{F:baseComp} and \cref{T:baseTable} illustrate our results. Our approach outperforms all baselines by a significant margin. This can be attributed to the new reward function that balances between exploring the environment and exploiting collected information. Both CAtNIPP variants perform worse than our method since they only focus on utility variance reduction. In \cref{T:baseTable}, the approaches using reinforcement learning require significantly less replanning time than non-learning methods, justifying the use of learning-based strategies. Our proposed approach is more compute-efficient than CAtNIPP (ts.), and almost as efficient as CAtNIPP (g.), which facilitates its deployment in real-world scenarios.

\cref{F:Qual_comp} qualitatively compares the paths planned by our approach and CAtNIPP g. for the ground truth environment illustrated in \cref{F:random}. The visualisations correspond to paths executed at $50\%$ of the budget. Our approach favours actions that discover more targets. This is because CAtNIPP considers a purely exploratory objective, assuming a continuous distribution of utility, which leads to re-observing the high-interest regions in an oscillatory manner. Our approach does not encounter this issue. Since we reduce both uncertainty and maximise the number of discovered targets using our reward function, we obtain a more widespread distribution of observations in the environment.

\subsection{Ablation Studies}
\label{ablation}

Next, we systematically study the impact of our dynamic graph action space and proposed reward function on the informativeness of the planned paths to demonstrate their benefits. Our test environment is the same as in~\Cref{Sec:BaseComp}. We compare our approach against CAtNIPP~\citep{cao2023catnipp}, which uses a static graph action space based on probabilistic roadmaps and a purely exploratory reward function.

\begin{table}[t]
    \vspace{0cm} \vfill
    \small
    \centering
    \caption{Graph structure ablation study. All methods use our proposed exploration-exploitation reward. The dynamic graph structure performs slightly better than the global representation of CAtNIPP.}%The cells indicate the mean values and standard deviation of the percentage of detected fruits.}

\begin{tabular}{l|p{1.75cm}<{\centering}|p{1.5cm}<{\centering}p{1.5cm}<{\centering}}
% \begin{tabular}
% {l|p{2cm}<{\centering}p{1cm}<{\centering}}
    %\toprule
        \multicolumn{1}{c|}{\multirow{1}{*}{Graph Structure}} & \multicolumn{1}{c|}{$\%$ targets} & \multicolumn{1}{c}{\edit{Time (s)}} \\
         % \multicolumn{1}{c|}{\multirow{1}{*}{Graph structure}} & \multicolumn{1}{c}{$\%$ targets} \\
         \midrule
         $K=10$ & $60.85 \pm 7.51$ & $1.40$\\% \pm 0.35$\\ 
         $K=20$ & $65.49 \pm 6.23$ & $1.58$\\% \pm 0.38$\\ 
         $K=30$ & $64.20 \pm 6.62$ & $1.99$\\% \pm 0.36$\\ 
         $K=40$ & $66.53 \pm 6.28$ & $2.75$\\% \pm 0.46$\\ 
         $K=50$ & $64.20 \pm 6.80$ & $3.80$\\% \pm 0.58$\\ 
         CAtNIPP~\citep{cao2023catnipp} & $63.27\pm 6.89$ & $1.28$\\% \pm 0.30$\\ 
     %    \bottomrule
    \end{tabular}
    \label{T:ablation_str}
    % \vspace{-0.15cm}
\end{table}

\begin{table}[t]
    \vspace{0cm} \vfill
    \small
    \centering
    \caption{Reward function ablation study. Both dynamic graph- and global graph-based policies trained on our reward function outperform those learned on a purely exploratory reward.}%The cells indicate mean values and standard deviation of the percentage of detected fruits.}
\begin{tabular}{l|p{2cm}<{\centering}p{1cm}<{\centering}}
%    \toprule
         \multicolumn{1}{c|}{\multirow{1}{*}{Model and reward function}} & \multicolumn{1}{c}{$\%$ targets} \\
         \midrule
         $K=20$, exploration & $54.75 \pm 7.66$\\ 
         $K=20$, our reward & $65.49 \pm 6.23$\\ 
         CAtNIPP~\citep{cao2023catnipp}, our reward & $63.27 \pm 6.89$\\ 
         CAtNIPP~\citep{cao2023catnipp}, exploration & $56.52 \pm 8.29$\\ 
%         \bottomrule
    \end{tabular}
    \label{T:ablation_rew}
    % \vspace{-0.65cm}
\end{table}

\textbf{Graph Structure}. \edit{To investigate the influence of the number of sampled waypoints $K$ on planning performance,} we compare our approach using dynamic graphs with $K~\in~\{10,\,20,\,30,\,40,\,50\}$ against CAtNIPP. \edit{In this section, w}e use our new reward function \edit{that combines both exploitation of obtained information and exploration of unknown regions, as} described \edit{in~\Cref{reward}, ensuring that the performance variations can be} attributed to the graph structure alone. \cref{T:ablation_str} summarises the results. We observe performance improvements from $K=10$ to $K=20$ \edit{and similar performance for $K \geq 20$ at the cost of increasing computation time}. The performance of our dynamic graph structure with $K\geq20$ is slightly better than the global graph of CAtNIPP. Hence, our dynamic graph can actively account for unknown obstacles while performing similarly to, or better than, the global graph structure.

\textbf{Reward Function}. Next, we investigate the effects of training a policy on our reward function against a purely exploratory reward function. We compare our dynamic graph-based approach trained on $K=20$ waypoints and $80$ action nodes against CAtNIPP trained on $K=200$ waypoints and $800$ action nodes. We tune the hyperparameters for best performance and consider two variants of each method with the different reward functions. For the purely exploratory reward, we set $\alpha=1$ and $\delta=0$ in \cref{reward}. \cref{T:ablation_rew} summarises the results. Both our dynamic graph and the CAtNIPP global graph trained on our new reward function outperform the corresponding policies trained using purely exploratory rewards. This confirms that learning from exploration rewards alone cannot guide the robot to adaptively focus on targets of interest since it incentivises actions that reduce overall utility variance. In contrast, our reward function incorporating both uncertainty reduction and targeted information gathering yields better \edit{target discovery} performance as it allows the policy to learn the trade-off between exploration and exploitation. The new reward function benefits both our dynamic graph and the global graph, showing its general applicability for different informative planning algorithms.

\subsection{Realistic Simulation}
\label{Sec:Col_avoid}

We demonstrate the applicability of our dynamic graph-based reinforcement learning approach with $K=20$ sampled waypoints in an orchard environment simulator created with Unreal Engine and AirSim. The Airsim simulator resembles real-world UAV dynamics, while Unreal Engine provides photorealistic imagery. Our apple orchard environment is bounded by a $95$\,m $\times 95$\,m $\times 18$\,m cuboid with $9$ trees arranged in a $3\times3$ array and a total of $225$ red apples at random locations on the trees as illustrated in~\cref{F:teaser}. We assume perfect localisation and use ground truth apple object \edit{discovery}. The \ac{UAV} moves at a maximum speed of $2$\,m/s.

\begin{figure}[!t]
\centering
\includegraphics[width=0.90\linewidth]{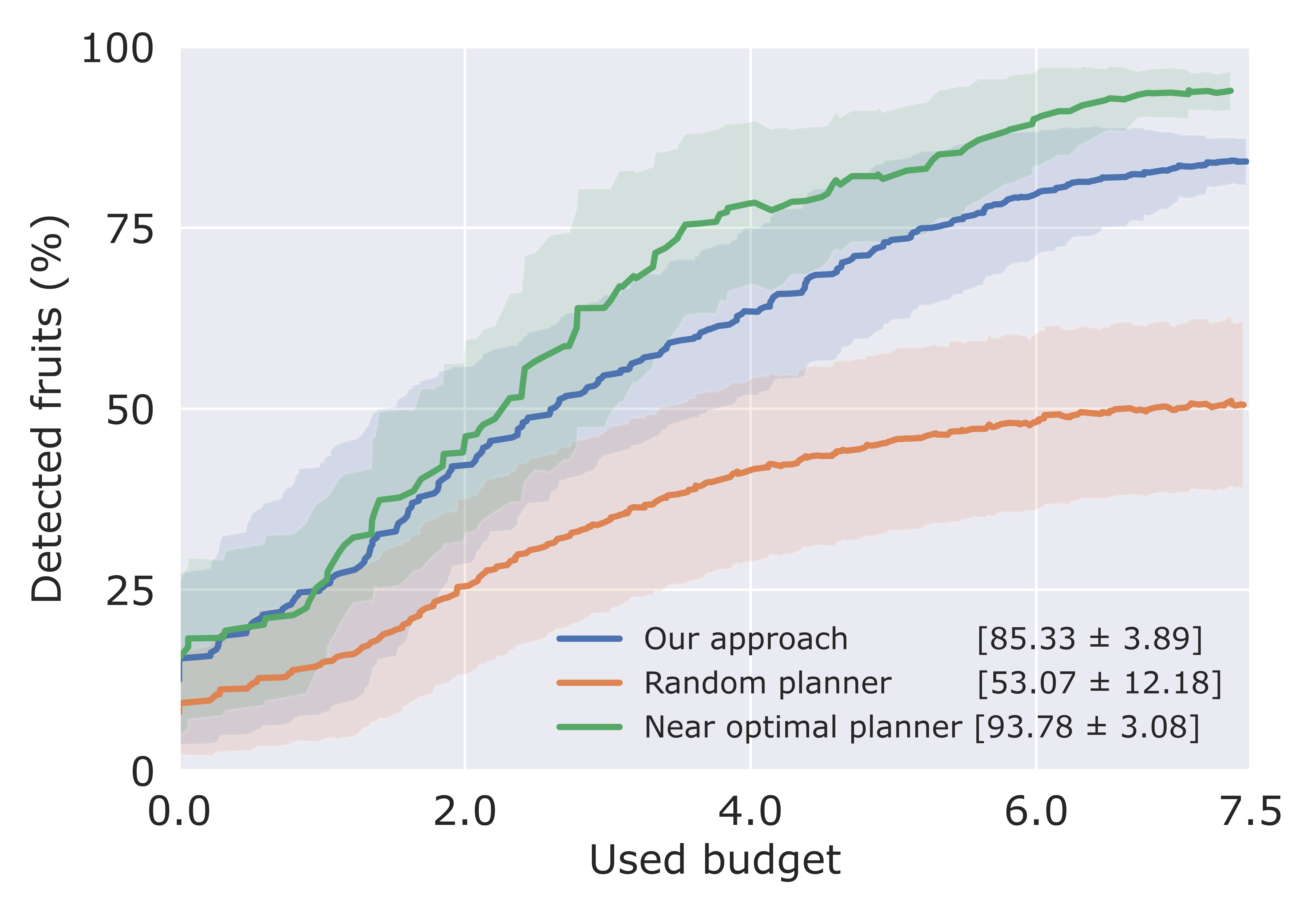}
% \vspace{-0.7cm}
\caption{Comparison of our reinforcement learning-based approach against baselines in a photorealistic fruit monitoring simulator. Solid lines indicate means over $10$ trials and shaded regions show standard deviations. Our approach performs almost as well as the near-optimal baseline, despite being trained in different environments and not relying on ground truth knowledge.}\small
\label{F:ue4Comp}
% \vspace{-0.6cm}
\end{figure}

We compare our approach trained in the synthetic simulation described in~\cref{Sec:implement} against (i)~a random planner over our $K=20$ dynamic graph and (ii)~a near-optimal planner to reflect performance upper bound using the metric of percentage discovered fruits. We record the coordinates of discovered fruits to ensure that the same fruit is not counted multiple times. We design the near-optimal planner to exploit ground truth information of the environment, such as tree coordinates, size, and best altitude, to generate a coverage-like path for observing maximum fruits. We run several instances of this planner and choose the three best-performing paths to compare against our approach. For our planner and the random planner, results are reported over $10$ trials in the environment with a mission budget of $7.5$ units.

\cref{F:ue4Comp} compares the three planners. Our approach outperforms non-informative random planning. The near-optimal planner performs best since it exploits ground truth knowledge to avoid viewpoint-dependent occlusions. However, our approach reaches similar performance without relying on any prior knowledge, making it suitable for unknown fruit distributions. \cref{F:teaser} visualises the path executed by our planner. These findings support the applicability of our method on a \ac{UAV} platform in a practical monitoring scenario.

% \begin{table}[t]
%     \vspace{0cm} \vfill
%     \small
%     \centering
%     \caption{\textbf{Comparison in Unreal Engine + Airsim environment.} The metrics used are $\%$ detected red apples with standard deviation in parentheses and budget utilised by the planner.}
% \begin{tabular}{l|p{2.0cm}<{\centering}|p{1.5cm}<{\centering}p{1cm}<{\centering}}
%     \toprule
%          \multicolumn{1}{c|}{\multirow{1}{*}{Planner Name}} & \multicolumn{1}{c|}{$\%$ Targets} & \multicolumn{1}{c}{Used budget} \\
%          \midrule
%          Near-Optimal Planner & $93.78(\pm3.08)$ & 7.36\\
%          Dynamic $k$-20 without $\mathbf{v}$ & $85.33(\pm3.89)$ & 7.40\\ 
%          Random Planner & $53.07(\pm12.18)$ & 7.50 \\
%          \bottomrule
%     \end{tabular}
%     \label{T:ue4Table}
% \vspace{-0.6cm}
% \end{table}

% Figure Labels: Use 8 point Times New Roman for Figure labels. Use words rather than symbols or abbreviations when writing Figure axis labels to avoid confusing the reader. Do not label axes only with units. Do not label axes with a ratio of quantities and units. For example, write ÒTemperature (K)Ó, not ÒTemperature/K.Ó

% A conclusion section is not required. Although a conclusion may review the main points of the paper, do not replicate the abstract as the conclusion. A conclusion might elaborate on the importance of the work or suggest applications and extensions. 

% \addtolength{\textheight}{-12cm}
\addtolength{\textheight}{0cm}   % This command serves to balance the column lengths

%%%%%%%%%%%%%%%%%%%%%%%%%%%%%%%%%%%%%%%%%%%%%%%%%%%%%%%%%%%%%%%%%%%%%%%%%%%%%%%%

%%%%%%%%%%%%%%%%%%%%%%%retaining%%%%%%%%%%%%%%%%%%%%%%%%%%%%%%%%%%%%%%%%%%%%%%%%%%%%%%%%%

%%%%%%%%%%%%%%%%%%%%%%%%%%%%%%%%%%%%%%%%%%%%%%%%%%%%%%%%%%%%%%%%%%%%%%%%%%%%%%%%
\section{Conclusion and Future Work}
We present a deep reinforcement learning approach for adaptively discovering targets of interest in unknown 3D environments. A key aspect of our method is a dynamic graph constructing a detailed environment representation to constrain planning in the robot's local region. We also present a new reward function enabling our learned policy to balance exploring the environment and exploiting obtained information. Our experimental results support our three claims: (i)~our approach outperforms the state-of-the-art learning-based approaches and non-learning baselines in environments unseen during training; (ii)~ our dynamic graph approach leads to performance on par, or better, than static global graph based state-of-the-art methods; (iii)~our new reward function outperforms a purely exploratory reward function. We validate our approach in a UAV-based fruit monitoring scenario to demonstrate its practical applicability. Future work \edit{includes developing advanced sampling strategies, considering dynamic obstacles,} and transferring our policy to a real robot under localisation and perception uncertainty. 
%and investigating the generalisability of our approach to different sensor setups and environments.

% \section{Future Work}


%%%%%%%%%%%%%%%%%%%%%%%%%%%%%%%%%%%%%%%%%%%%%%%%%%%%%%%%%%%%%%%%%%%%%%%%%%%%%%%%

% References are important to the reader; therefore, each citation must be complete and correct. If at all possible, references should be commonly available publications.

\bibliographystyle{IEEEtranN}
\footnotesize
\bibliography{root}

\end{document}